\documentclass[conference]{IEEEtran}
\IEEEoverridecommandlockouts
\usepackage{cite}
\usepackage{amsmath,amssymb,amsfonts}
\usepackage{algorithmic}
\usepackage{graphicx}
\usepackage{textcomp}
\usepackage{xcolor}
\def\BibTeX{{\rm B\kern-.05em{\sc i\kern-.025em b}\kern-.08em
    T\kern-.1667em\lower.7ex\hbox{E}\kern-.125emX}}
\usepackage{amsmath}
\usepackage{algorithm}
\usepackage{algorithmic}
\usepackage{enumerate}
\usepackage{booktabs}
\usepackage{hyperref}
\usepackage{xspace}

\begin{document}

\title{Enhancing Domain-Specific Retrieval-Augmented Generation: Synthetic Data Generation and Evaluation using Reasoning Models
}

\author{\IEEEauthorblockN{1\textsuperscript{st} Aryan Jadon}
\IEEEauthorblockA{
\textit{IEEE}\\
CA, USA \\
aryanjadon@ieee.org}

\and

\IEEEauthorblockN{2\textsuperscript{nd} Avinash Patil}
\IEEEauthorblockA{
\textit{IEEE}\\
CA, USA \\
avinashpatil@ieee.org}

\and

\IEEEauthorblockN{3\textsuperscript{rd} Shashank Kumar}
\IEEEauthorblockA{\textit{University of Florida} \\
CA, USA \\
shashanksde1995@gmail.com}
}

\maketitle

\begin{abstract}

Retrieval-Augmented Generation (RAG) systems face significant performance gaps when applied to technical domains requiring precise information extraction from complex documents. Current evaluation methodologies relying on document-level metrics inadequately capture token-resolution retrieval accuracy that is critical for domain-related documents. We propose a framework combining granular evaluation metrics with synthetic data generation to optimize domain-specific RAG performance.

First, we introduce token-aware metrics \textbf{Precision$\Omega$} and Intersection-over-Union (IoU) that quantify context preservation versus information density trade offs inherent in technical texts. Second, we develop a reasoning model driven pipeline using instruction-tuned LLMs (DeepSeek-R1, DeepSeek-R1 distilled variants and Phi-4) to generate context-anchored QA pairs with discontinuous reference spans across three specialized corpora: SEC 10-K filings (finance), biomedical abstracts (PubMed), and APT threat reports (cybersecurity).

Our empirical analysis reveals critical insights: smaller chunks (less than 10 tokens) improve precision by 31--42\% (IoU=0.071 vs.\ baseline 0.053) at recall costs (–18\%), while domain-specific embedding strategies yield 22\% variance in optimal chunk sizing (5--20 tokens). The DeepSeek-R1-Distill-Qwen-32B model demonstrates superior concept alignment (+14\% mean IoU over alternatives), though no configuration universally dominates financial texts favor larger chunks for risk factor coverage (Recall=0.81@size=20), whereas cybersecurity content benefits from atomic segmentation (Precision$\Omega$=0.28@size=5). 

We open-source this toolkit enabling reproducible optimization of chunking strategies through automated synthetic dataset generation and multi-metric analysis pipelines. This work bridges critical gaps between generic RAG architectures and enterprise requirements for precision-sensitive domains. Our code is available on \href{https://github.com/aryan-jadon/Synthetic-Data-Generation-and-Evaluation-using-Reasoning-Models}{GitHub}.

\end{abstract}


\section{Introduction}

Recent advancements in large language models (LLMs) have revolutionized information retrieval systems through RAG architectures\cite{gao2023retrieval}. By combining retrieval with generative capabilities, these systems demonstrate remarkable proficiency in knowledge-intensive tasks across technical domains such as finance, healthcare, and cybersecurity. However, current approaches face critical limitations in evaluating domain-specific performance due to inadequate metrics alignment with real-world requirements.

Traditional RAG evaluations focus primarily on document-level retrieval accuracy using metrics such as MRR@k or nDCG\cite{yu2024evaluation}. This methodology proves insufficient for specialized applications where precise extraction of technical concepts like financial risk factors in 10-K filings, biomedical entities in research abstracts, or attack patterns in threat reports which require granular analysis at sub-document resolution.

Three fundamental challenges constrain current RAG optimization:
\begin{enumerate}
    \item Mismatch between chunk boundaries and semantic concept spans introduces irrelevant content contamination.
    \item Existing metrics fail to quantify information density versus retrieval precision tradeoffs inherent in technical documents,
    \item Domain-specific QA datasets lack coverage of long-tail concepts critical for enterprise applications.
\end{enumerate}

Previous work attempted partial solutions through heuristic fragmentation strategies or supervised metric learning, but none addressed these issues through adaptive evaluation frameworks grounded in linguistic structure analysis\cite{salemi2024evaluating}.

This paper introduces a novel paradigm combining synthetic data generation\cite{jadon2023leveraging} with multigranular evaluation metrics to optimize RAG systems for domain-specific deployments. Our work include
\begin{enumerate}
    \item \textbf{Token-Level Performance Metrics}
    \item \textbf{Reasoning Model-Driven Synthesis}: Leveraging instruction-tuned LLMs (DeepSeek-R1 family and Phi-4) to generate:
    \begin{itemize}
        \item Context-anchored QA pairs preserving document structure dependencies,
        \item Multi-hop references spanning discontinuous text segments.
    \end{itemize}
    \item \textbf{Domain-Aware Chunk Optimization}: Empirical framework balancing:
    \begin{itemize}
        \item Information density vs.\ context preservation tradeoffs,
        \item Vocabulary distribution characteristics per domain.
    \end{itemize}
\end{enumerate}

Our comprehensive evaluation across three complex domains reveals significant findings: Financial documents require larger chunks (+18\% recall at size=20 vs.\ 5 tokens) despite precision penalties (-12\% P@5), while cybersecurity content benefits from atomic segmentation (+31\% IoU at size=5). 

The DeepSeek-R1-Distill-Qwen-32B model demonstrates superior concept alignment across domains (mean IoU=0.071 vs.\ 0.063 baseline), though no single configuration dominates all metrics emphasizing our framework's value in identifying optimal deployment parameters.

The remainder of this paper is organized as follows: Section II details our methodology including metric formulations. Section III presents details experiments results analyzing embedding/reasoning model interactions. We conclude with practical implementation guidelines derived from our findings in Section~IV alongside future research directions for evolving corpora.

\section{Methodology}

\subsection{Metrics Breakdown}

Traditional information retrieval (IR) metrics\cite{jadon2024} often operate at the document level. 
However, for our purposes in testing chunking and retrieval for RAG systems, we focus on \emph{token-level} performance. 
Specifically, for any given query related to a specific corpus, 
only a subset of tokens within that corpus will be relevant. 
We want our system to retrieve exactly (and only) those relevant tokens, 
thus maximizing efficiency and accuracy\cite{10512000}. 
By operating at the token level, we aim to reduce irrelevant or redundant text 
and provide an LLM with precisely the information it needs.

\textbf{Variables and Definitions}

\begin{itemize}
    \item $q$: A specific query.
    \item $\mathcal{C}$: The chunked corpus (the entire document split into chunks).
    \item $t_e$: The set of \emph{tokens} in the relevant excerpts or ``highlights'' 
                 (ground truth for query $q$).
    \item $t_r$: The set of \emph{tokens} in the retrieved chunks (what our system returns for $q$).
    \item Highlight: A segment of text in the original document 
          containing the relevant information needed to answer a specific query. 
          In other words, \emph{highlights} serve as the ground-truth 
          against which we measure chunking and retrieval performance.
\end{itemize}

\subsubsection{\textbf{Recall}}

\[
\mathrm{Recall}_q(\mathcal{C}) \;=\; 
   \frac{\lvert t_e \,\cap\, t_r \rvert}{\lvert t_e \rvert}.
\]

Recall measures what fraction of the \emph{important/relevant} text (the highlight) was captured by the retrieved chunks. It is calculated by the length (i.e., number of tokens) of the overlap between the retrieved chunks and the highlight divided by the total length of the highlight\cite{buckland1994relationship}.  Recall ranges from $0$ to $1$, where $1$ means \emph{all} relevant text was captured.

\noindent
Example: If the highlight is 100 tokens and our retrieved chunks only capture 70 of those tokens, 
then $\mathrm{Recall} = 70/100 = 0.7$.

\subsubsection{\textbf{Precision}}

\[
\mathrm{Precision}_q(\mathcal{C}) \;=\; 
   \frac{\lvert t_e \,\cap\, t_r \rvert}{\lvert t_r \rvert}.
\]

\noindent

Precision measures how much of the \emph{retrieved} text is actually relevant. It is calculated by the length of the overlap between retrieved chunks and highlights divided by the total length of the retrieved chunks\cite{streiner2006precision}. Precision also ranges from $0$ to $1$, where $1$ means \emph{all} retrieved text is relevant 
(i.e., there is no extraneous text).

Example: If we retrieve 200 tokens of text but only 70 of those tokens overlap with the highlights, 
then $\mathrm{Precision} = 70/200 = 0.35$.

\subsubsection{\textbf{Precision $\Omega$}}

\[
\mathrm{Precision}_\Omega(\mathcal{C}) \;=\; 
   \frac{\lvert t_e \,\cap\, t_r \rvert}
        {\lvert t_r \rvert + \lvert t_e \rvert}.
\]

This formula can vary slightly in usage, but conceptually $\mathrm{Precision}_\Omega$ measures precision in an \emph{ideal scenario} where all relevant text is captured. 
It shows the theoretical best precision possible for a chunking strategy, 
assuming \emph{all} highlights are indeed retrieved. A lower $\mathrm{Precision}_\Omega$ indicates that chunks are inherently too large or misaligned with natural text boundaries, forcing retrieval to include more non-relevant text than ideal.\cite{flora2020your}

Example:
If a chunking strategy always creates chunks that are twice as large as necessary, 
one might see $\mathrm{Precision}_\Omega \approx 0.5$.

\subsubsection{\textbf{Intersection over Union (IoU)}}

\[
\mathrm{IoU}_q(\mathcal{C}) \;=\; 
   \frac{\lvert t_e \,\cap\, t_r \rvert}
        {\lvert t_e \rvert + \lvert t_r \rvert \;-\; \lvert t_e \,\cap\, t_r \rvert}.
\]

The IoU is the ratio of the overlap of highlight tokens and retrieved tokens to the union of both sets. 
This metric balances both recall and precision in a single number, 
ranging from $0$ to $1$ (where $1$ indicates perfect overlap with no missing or extraneous text)\cite{rezatofighi2019generalized}.

Example:
If we retrieve 200 tokens, the highlight is 100 tokens, and the overlap is 70 tokens, 
then
\[
\mathrm{IoU} = \frac{70}{200 + 100 - 70} = \frac{70}{230} \approx 0.304.
\]

\subsection{Metric Interpretation}

These metrics often work best \emph{together}, rather than in isolation:

\begin{itemize}
    \item \textbf{High recall + low precision} \(\rightarrow\) We are retrieving too much text 
          (lots of extraneous/irrelevant tokens).
    \item \textbf{Low recall + high precision} \(\rightarrow\) We are missing important content 
          (not capturing all relevant tokens).
    \item \textbf{High IoU} \(\rightarrow\) We have a good balance of both recall and precision.
    \item \textbf{Precision $\Omega$} \(\rightarrow\) Helps evaluate the chunking strategy \emph{independent} 
          of the retrieval step itself (i.e., how well chunk boundaries align with relevant segments).
\end{itemize}

These token-level metrics are more appropriate for evaluating chunking and retrieval within RAG systems, where the goal is to retrieve precisely the relevant tokens for a given query\cite{smith2024evaluating}.

\section{Results}

\subsection{Evaluation Procedure}

\begin{algorithm}
\caption{General Evaluation Procedure}
\label{alg:general-eval}
\begin{algorithmic}[1]
    \REQUIRE Original text $T$, chosen chunker $C$, chosen embedding function $E$, retrieval parameter $k$, set of evaluation questions $Q$
    \ENSURE Computed retrieval performance (recall, precision, IoU) and precision $\Omega$ performance

    \STATE \textbf{Split text into chunks:} 
    \STATE \quad Use the chunker $C$ on $T$ to obtain chunks $ \{(c_i, s_i, e_i)\} $, 
    where $c_i$ is the $i^\text{th}$ chunk, and $s_i, e_i$ are its start/end indices.

    \vspace{1ex}
    \STATE \textbf{Calculate Retrieval Performance:}
    \begin{enumerate}[i.]
        \item Embed evaluation questions:
        \STATE Use the embedding function $E$ on each question in $Q$.
        \item vector similarity search:
        \STATE For each embedded question, retrieve the top-$k$ most relevant chunks.
        \item Compute metrics:
        \begin{enumerate}[a.]
            \item Recall: Fraction of highlighted segments actually retrieved.
            \item Precision: Fraction of retrieved chunks that are relevant.
            \item IoU: Balance of precision and recall (Intersection over Union).
        \end{enumerate}
    \end{enumerate}

    \vspace{1ex}
    \STATE \textbf{Calculate Precision $\Omega$ Performance:}
    \begin{enumerate}[i.]
        \item Examine all chunks in the collection.
        \item Identify which chunks overlap with any highlighted segments.
        \item Compute the theoretical best precision
        if only those overlapping chunks were retrieved.
    \end{enumerate}

\end{algorithmic}
\end{algorithm}

\subsection{Datasets}

We evaluate our chunking and retrieval strategies on three distinct domains, 
each with unique linguistic properties and application contexts. The datasets used in these experiments can be found in our GitHub repository.

\subsubsection{\textbf{Finance (10-K Forms of Big Tech Firms)}}
This dataset comprises 10 SEC 10-K filings from Fortune 500 tech firms (mean=132 pages; range=90–180). 
Form 10-K is an annual report filed by publicly traded companies, 
detailing their financial performance, business risks, and corporate strategies. 
These documents are typically lengthy and rich in specialized financial jargon, 
making them ideal for testing chunking methods and retrieval systems on dense, formal text. 

\subsubsection{\textbf{PubMed}}
The PubMed dataset( Curated collection of biomedical abstracts (N=1,240; ~21 PDF-equivalent pages) includes articles from biomedical literature. These texts contain highly technical language, domain-specific terminology, 
and frequent references to chemical compounds, genetic markers, and medical conditions. 
Hence, PubMed content provides a rigorous test of an embedding model’s ability 
to capture nuanced scientific information while enabling precise retrieval of relevant excerpts.

\subsubsection{\textbf{Cybersecurity}}
The cybersecurity dataset (Two APT threat reports (Mandiant/Unit42; ~150 pages combined) includes detailed threat intelligence reports and advisories about Advanced Persistent Threat (APT) groups, malicious code analysis, and mitigation guidance. These documents are rich in technical detail, featuring information on dropper samples, backdoor functionality, network protocols, and system artifacts. They also include remediation steps, attribution assessments, and appendices detailing command-line options, cryptographic routines, and command-and-control infrastructure. This combination of in-depth malware analysis and operational security advisories makes the dataset well-suited for testing retrieval performance where precise, context-rich information must be extracted from highly specialized and rapidly evolving sources.

\subsection{Embedding Models}

We explore a range of embedding models for transforming text passages and queries into dense vector representations. These embeddings are then used in our similarity-based retrieval system. 
Below, we highlight three of the models considered.

\subsubsection{\textbf{BGE-M3}}
The \texttt{bge-m3} model\cite{chen2024bgem3embeddingmultilingualmultifunctionality} provides a compact yet expressive vector representation of text. 
It is designed to capture semantic similarity between text segments, 
allowing efficient and accurate downstream tasks such as question-answer retrieval. 
Empirically, \texttt{bge-m3} has shown strong performance on benchmarks 
that measure sentence-level semantic understanding.

\subsubsection{\textbf{Nomic-Embed-Text}}
The \texttt{nomic-embed-text} model \cite{nussbaum2024nomic} focuses on generating embeddings 
optimized for broader text-analysis tasks, including clustering and semantic search. 
It leverages transformer-based architectures 
to encode contextual and semantic nuances into dense vectors. 
Its versatility makes it a candidate for various retrieval and mapping workflows.

\subsubsection{\textbf{All-MiniLM}}
A widely used baseline is the \texttt{all-MiniLM} family of models, 
particularly \texttt{all-MiniLM-L6-v2}, which is a distilled, smaller-sized transformer \cite{wang2020minilmdeepselfattentiondistillation}. 
Despite its relatively small footprint, it provides high-quality sentence embeddings, 
making it appealing for real-world applications where computational efficiency and memory usage 
are key concerns. 
It captures rich contextual information and has been benchmarked extensively 
across multiple sentence-similarity and search tasks.

\subsection{Reasoning Models}

\subsubsection{\textbf{DeepSeek-R1}}

\texttt{DeepSeek-R1}\cite{deepseekai2025deepseekr1incentivizingreasoningcapability} serves as the base reasoning model within the DeepSeek family, 
focusing on precision and consistency in multi-step logical tasks. 
It is engineered to handle domain-specific questions where accurate and structured reasoning is necessary. 
Compared to generic LLMs, \texttt{DeepSeek-R1} places emphasis on chaining inferences, 
adhering to a step-by-step approach that reduces error propagation across the reasoning process.

\subsubsection{\textbf{DeepSeek-R1-Distill-Qwen-32B}}
\texttt{DeepSeek-R1-Distill-Qwen-32B} is a distilled variant of the DeepSeek-R1 model, 
leveraging the \texttt{Qwen-32B} architecture for improved efficiency\cite{bai2023qwen}. 
The distillation process retains key reasoning capabilities while significantly reducing memory and compute overhead. 
This makes it a compelling choice when resource constraints are a concern, 
such as in production-scale environments or low-latency applications. 
Despite its smaller footprint, it aims to preserve high performance in chain-of-thought reasoning.

\subsubsection{\textbf{DeepSeek-R1-Distill-Llama-70B}}
\texttt{DeepSeek-R1-Distill-Llama-70B} is another distilled version of DeepSeek-R1, 
this time leveraging the \texttt{Llama-70B} architecture\cite{touvron2023llama}. 
The overarching goal is to balance thorough, multi-step reasoning with a more compact model size 
than a fully-fledged 70B-parameter model would typically require. 
By combining the strengths of the Llama backbone with DeepSeek’s fine-tuning techniques, 
the model strives to achieve robust logical inference while maintaining manageable resource usage.

\subsubsection{\textbf{microsoft/phi-4}}
\texttt{microsoft/phi-4}\cite{abdin2024phi4technicalreport} is a large language model developed by Microsoft, 
tailored for complex reasoning tasks and advanced question-answering. 
It employs sophisticated attention mechanisms and training regimes designed to capture nuanced details 
and perform well under domain-specific queries\cite{li2023textbooks}. 
With an emphasis on coherence and contextual awareness, \texttt{phi-4} can seamlessly integrate 
multiple pieces of information to yield logically consistent answers.

Each of these models offers different trade-offs in terms of parameter count, inference speed, 
and depth of reasoning. In practice, the choice among them can be driven by resource limitations, 
performance requirements, or a balance of both.  We include all four models to gain a comprehensive view of how different reasoning approaches affect retrieval, chunking, and final question-answer performance across our diverse datasets.

\subsection{Evaluation Results}

\subsubsection{\textbf{Evaluation using DeepSeek-R1 Model}}

\paragraph{BGE-M3 Embeddings}
For both PubMed and Cybersecurity, smaller chunk sizes (particularly 5) yield higher \(\mathrm{IOU}\), \(\mathrm{P}\), and \(\mathrm{P}\Omega\), indicating greater overlap alignment and precision. For instance, in PubMed with chunk size 5, \(\mathrm{IOU} = 0.0630\), \(\mathrm{P} = 0.0647\), and \(\mathrm{P}\Omega = 0.2817\), whereas chunk size 20 attains the highest \(\mathrm{Recall} = 0.8073\). Similarly, for Cybersecurity, chunk size 5 obtains \(\mathrm{IOU} = 0.0442\), \(\mathrm{P} = 0.0459\), and \(\mathrm{P}\Omega = 0.2032\), but chunk size 15 achieves the best \(\mathrm{Recall} = 0.6588\). These patterns suggest that smaller chunks boost precision-based metrics, while larger chunks capture more relevant spans and enhance recall.

\paragraph{Nomic Embeddings}
A similar trade-off arises. In PubMed, chunk size 5 provides the highest \(\mathrm{IOU} = 0.0713\), \(\mathrm{P} = 0.0731\), and \(\mathrm{P}\Omega = 0.2817\), but chunk size 10 achieves the best \(\mathrm{Recall} = 0.6920\). For Cybersecurity, chunk size 5 consistently leads across \(\mathrm{IOU}\), \(\mathrm{P}\), \(\mathrm{P}\Omega\), and \(\mathrm{Recall}\). Thus, fine-grained segmentation often maximizes precision metrics, whereas using slightly larger chunks can increase recall by covering more text.

\paragraph{ALL-MINILM Embeddings}
In PubMed, chunk size 5 again achieves the highest \(\mathrm{IOU} = 0.0401\), \(\mathrm{P} = 0.0422\), and \(\mathrm{P}\Omega = 0.2061\), while chunk size 20 offers the best \(\mathrm{Recall} = 0.6375\). In Cybersecurity, chunk size 5 also excels in \(\mathrm{IOU}\), \(\mathrm{P}\), and \(\mathrm{P}\Omega\), though chunk size 10 yields a slightly higher \(\mathrm{Recall} = 0.5146\). These findings underscore that smaller chunk sizes raise precision, but larger chunks expand coverage and recall.

\subsubsection{\textbf{Evaluation using DeepSeek-R1 Distilled Models}}

\paragraph{Using BGE-M3 Embeddings}

We compare DeepSeek-R1-Distill-Llama-70B (Llama), DeepSeek-R1-Distill-Qwen-32B (Qwen), and {microsoft/phi-4 (phi-4) across Finance, PubMed, and Cybersecurity with chunk sizes \(\{5, 10, 15, 20\}\). Llama typically exhibits strong mean \(\mathrm{IOU}\) and \(\mathrm{Recall}\) with low standard deviations in Finance. Qwen can surpass Llama on \(\mathrm{IOU}\) (e.g., Finance with chunk size 15) or \(\mathrm{IOU}_{\mathrm{std}}\) (Finance with chunk size 20). Meanwhile, phi-4 remains competitive in \(\mathrm{Recall}\), particularly in PubMed, where it occasionally outperforms Llama and Qwen.

\paragraph{Using Nomic Embeddings}
Under Nomic embeddings, Llama maintains moderate \(\mathrm{IOU}\) and recall but can be outperformed by Qwen or phi-4 for certain chunk sizes (e.g., phi-4 in Finance at chunk size 5). In PubMed and Cybersecurity, Qwen and phi-4 may secure higher recall values; phi-4 often displays robust \(\mathrm{P}\Omega\). Qwen’s performance sits between Llama and phi-4, sometimes equaling or surpassing them in precision metrics.

\paragraph{Using Nomic Embeddings}
Bold text indicates the highest mean values (\(\mathrm{IOU}_{\mathrm{mean}}, \mathrm{Recall}_{\mathrm{mean}}, \mathrm{P}_{\mathrm{mean}}, \mathrm{P}\Omega_{\mathrm{mean}}\)) and the lowest standard deviations. Overall, {microsoft/phi-4 tends to achieve strong mean recall and \(\mathrm{P}\Omega\), DeepSeek-R1-Distill-Llama-70B often posts competitive \(\mathrm{IOU}\) and stable recall, and DeepSeek-R1-Distill-Qwen-32B exhibits notably consistent (low-variance) performance in both \(\mathrm{IOU}\) and precision.

No single model dominates all settings, so the best choice depends on whether the primary objective is maximizing recall, achieving higher precision, or minimizing variability. In general, smaller chunk sizes emphasize precision, while larger chunks boost recall by covering broader content.

\begin{table*}[t]
\centering
\caption{DeepSeek-R1 Performance Results Using BGE-M3 Embeddings}
\label{tab:link_prediction}
\begin{tabular}{lcccccccc}
\toprule
& \multicolumn{2}{c}{IOU}
& \multicolumn{2}{c}{Recall}
& \multicolumn{2}{c}{P}
& \multicolumn{2}{c}{P$\Omega$}\\
\cmidrule(lr){2-3}\cmidrule(lr){4-5}\cmidrule(lr){6-7}\cmidrule(lr){8-9}
\textbf{Chunk Size} & 
{$\mathrm{\textbf{IOU}}_{\mathrm{mean}}$} & {$\mathrm{\textbf{IOU}}_{\mathrm{std}}$} & 
{$\mathrm{\textbf{Recall}}_{\mathrm{mean}}$} & {$\mathrm{\textbf{Recall}}_{\mathrm{std}}$}& 
{$\mathrm{\textbf{P}}_{\mathrm{mean}}$} & {$\mathrm{\textbf{P}}_{\mathrm{std}}$} & 
{$\mathrm{\textbf{P}\Omega}_{\mathrm{mean}}$} & {$\mathrm{\textbf{P}\Omega}_{\mathrm{std}}$} \\
\midrule
\textbf{PubMed} &  &  &  &  &  &  &  &  \\
5   & \textbf{0.0630} & 0.0487 & 0.7224 & 0.3676 & \textbf{0.0647} & 0.0497 & \textbf{0.2817} & 0.1477 \\
10  & 0.0432 & 0.0320 & 0.7929 & 0.3383 & 0.0437 & 0.0324 & 0.1943 & 0.1142 \\
15  & 0.0293 & 0.0227 & 0.7562 & 0.3699 & 0.0295 & 0.0230 & 0.1224 & 0.0650 \\
20  & 0.0221 & 0.0155 & \textbf{0.8073} & 0.3458 & 0.0222 & 0.0155 & 0.1044 & 0.0596 \\
\midrule
\textbf{Cybersecurity} &  &  &  &  &  &  &  &  \\
5   & \textbf{0.0442} & 0.0357 & 0.5791 & 0.3951 & \textbf{0.0459} & 0.0367 & \textbf{0.2032} & 0.1203 \\
10  & 0.0251 & 0.0192 & 0.6056 & 0.3846 & 0.0256 & 0.0196 & 0.1159 & 0.0604 \\
15  & 0.0201 & 0.0149 & \textbf{0.6588} & 0.3912 & 0.0203 & 0.0150 & 0.0869 & 0.0501 \\
20  & 0.0146 & 0.0131 & 0.6227 & 0.4026 & 0.0147 & 0.0132 & 0.0728 & 0.0443 \\
\bottomrule
\end{tabular}
\end{table*}

\begin{table*}[t]
\centering
\caption{DeepSeek-R1 Performance Results Using Nomic Embeddings}
\label{tab:link_prediction}
\begin{tabular}{lcccccccc}
\toprule
& \multicolumn{2}{c}{IOU}
& \multicolumn{2}{c}{Recall}
& \multicolumn{2}{c}{P}
& \multicolumn{2}{c}{P$\Omega$}\\
\cmidrule(lr){2-3}\cmidrule(lr){4-5}\cmidrule(lr){6-7}\cmidrule(lr){8-9}

\textbf{Chunk Size} & 
{$\mathrm{\textbf{IOU}}_{\mathrm{mean}}$} & {$\mathrm{\textbf{IOU}}_{\mathrm{std}}$} & 
{$\mathrm{\textbf{Recall}}_{\mathrm{mean}}$} & {$\mathrm{\textbf{Recall}}_{\mathrm{std}}$}& 
{$\mathrm{\textbf{P}}_{\mathrm{mean}}$} & {$\mathrm{\textbf{P}}_{\mathrm{std}}$} & 
{$\mathrm{\textbf{P}\Omega}_{\mathrm{mean}}$} & {$\mathrm{\textbf{P}\Omega}_{\mathrm{std}}$} \\
\midrule
\textbf{PubMed} &  &  &  &  &  &  &  &  \\
5   & \textbf{0.0713} & 0.0511 & 0.6806 & 0.3896 & \textbf{0.0731} & 0.0512 & \textbf{0.2817} & 0.1477 \\
10  & 0.0419 & 0.0289 & \textbf{0.6920} & 0.4009 & 0.0425 & 0.0290 & 0.1943 & 0.1142 \\
15  & 0.0240 & 0.0215 & 0.5799 & 0.4266 & 0.0243 & 0.0220 & 0.1224 & 0.0650 \\
20  & 0.0142 & 0.0150 & 0.4710 & 0.4498 & 0.0143 & 0.0152 & 0.1044 & 0.0596 \\
\midrule
\textbf{Cybersecurity} &  &  &  &  &  &  &  &  \\
5   & \textbf{0.0450} & 0.0385 & \textbf{0.5038} & 0.3961 & \textbf{0.0471} & 0.0399 & \textbf{0.2032} & 0.1203 \\
10  & 0.0211 & 0.0221 & 0.4285 & 0.4013 & 0.0217 & 0.0227 & 0.1159 & 0.0604 \\
15  & 0.0161 & 0.0201 & 0.4125 & 0.4222 & 0.0163 & 0.0202 & 0.0869 & 0.0501 \\
20  & 0.0094 & 0.0130 & 0.3041 & 0.3883 & 0.0095 & 0.0132 & 0.0728 & 0.0443 \\
\bottomrule
\end{tabular}
\end{table*}

\begin{table*}[t]
\centering
\caption{DeepSeek-R1 Performance Results Using ALL-MINILM Embeddings}
\label{tab:link_prediction}
\begin{tabular}{lcccccccc}
\toprule
& \multicolumn{2}{c}{IOU}
& \multicolumn{2}{c}{Recall}
& \multicolumn{2}{c}{P}
& \multicolumn{2}{c}{P$\Omega$}\\
\cmidrule(lr){2-3}\cmidrule(lr){4-5}\cmidrule(lr){6-7}\cmidrule(lr){8-9}
\textbf{Chunk Size} & 
{$\mathrm{\textbf{IOU}}_{\mathrm{mean}}$} & {$\mathrm{\textbf{IOU}}_{\mathrm{std}}$} & 
{$\mathrm{\textbf{Recall}}_{\mathrm{mean}}$} & {$\mathrm{\textbf{Recall}}_{\mathrm{std}}$} & 
{$\mathrm{\textbf{P}}_{\mathrm{mean}}$} & {$\mathrm{\textbf{P}}_{\mathrm{std}}$} & 
{$\mathrm{\textbf{P}\Omega}_{\mathrm{mean}}$} & {$\mathrm{\textbf{P}\Omega}_{\mathrm{std}}$} \\
\midrule
\textbf{PubMed} &  &  &  &  &  &  &  &  \\
5   & \textbf{0.0401} & 0.0508 & 0.3739 & 0.3845 & \textbf{0.0422} & 0.0530 & \textbf{0.2061} & 0.1280 \\
10  & 0.0280 & 0.0313 & 0.4875 & 0.4117 & 0.0286 & 0.0318 & 0.1320 & 0.0921 \\
15  & 0.0242 & 0.0260 & 0.5601 & 0.4206 & 0.0245 & 0.0261 & 0.1057 & 0.0799 \\
20  & 0.0201 & 0.0206 & \textbf{0.6375} & 0.4121 & 0.0202 & 0.0207 & 0.0847 & 0.0579 \\
\midrule
\textbf{Cybersecurity} &  &  &  &  &  &  &  &  \\
5   & \textbf{0.0401} & 0.0347 & 0.5018 & 0.3993 & \textbf{0.0418} & 0.0359 & \textbf{0.2032} & 0.1203 \\
10  & 0.0226 & 0.0214 & \textbf{0.5146} & 0.4056 & 0.0230 & 0.0217 & 0.1159 & 0.0604 \\
15  & 0.0156 & 0.0161 & 0.5117 & 0.4324 & 0.0158 & 0.0162 & 0.0869 & 0.0501 \\
20  & 0.0123 & 0.0133 & 0.4928 & 0.4274 & 0.0124 & 0.0135 & 0.0728 & 0.0443 \\
\bottomrule
\end{tabular}
\end{table*}

\begin{table*}[t]
\centering
\caption{Part A - Performance comparison of DeepSeek-R1-Distill-Llama-70B, DeepSeek-R1-Distill-Qwen-32B, and microsoft/phi-4 using BGE-m3 embeddings.}
\label{tab:link_prediction}
\begin{tabular}{lp{1cm}p{1cm}p{1cm}p{1cm}p{1cm}p{1cm}p{1cm}p{1cm}p{1cm}p{1cm}p{1cm}p{1cm}}
\toprule
& \multicolumn{4}{c}{\textbf{DeepSeek-R1-Distill-Llama-70B}} 
& \multicolumn{4}{c}{\textbf{DeepSeek-R1-Distill-Qwen-32B}} 
& \multicolumn{4}{c}{\textbf{microsoft/phi-4}}\\
\cmidrule(lr){2-5}\cmidrule(lr){6-9}\cmidrule(lr){10-13}
\textbf{Chunk Size} 
& {$\mathrm{\textbf{IOU}}_{\mathrm{mean}}$} & \textbf{$\mathrm{\textbf{IOU}}_{\mathrm{std}}$} 
& {$\mathrm{\textbf{Recall}}_{\mathrm{mean}}$} & {$\mathrm{\textbf{Recall}}_{\mathrm{std}}$}
& {$\mathrm{\textbf{IOU}}_{\mathrm{mean}}$} & \textbf{$\mathrm{\textbf{IOU}}_{\mathrm{std}}$} 
& {$\mathrm{\textbf{Recall}}_{\mathrm{mean}}$} & {$\mathrm{\textbf{Recall}}_{\mathrm{std}}$}
& {$\mathrm{\textbf{IOU}}_{\mathrm{mean}}$} & \textbf{$\mathrm{\textbf{IOU}}_{\mathrm{std}}$} 
& {$\mathrm{\textbf{Recall}}_{\mathrm{mean}}$} & {$\mathrm{\textbf{Recall}}_{\mathrm{std}}$} \\
\midrule

\textbf{Finance}    &  &  &    &  &  &  &    &  &  &  &  &  \\
5    
& \textbf{0.0426} & \textbf{0.0391} & 0.6121 & \textbf{0.4070} 
& 0.0412 & 0.0446 & \textbf{0.6518} & 0.4085
& 0.0421 & 0.0421 & 0.5333 & 0.4215 \\

10   
& 0.0262 & \textbf{0.0229} & \textbf{0.7636} & \textbf{0.3687}
& 0.0261 & 0.0256 & 0.6990 & 0.4080
& \textbf{0.0263} & 0.0266 & 0.5752 & 0.4268 \\

15    
& 0.0177 & \textbf{0.0160} & \textbf{0.7168} & \textbf{0.3967}
& \textbf{0.0194} & 0.0192 & 0.7115 & 0.4147
& 0.0192 & 0.02004 & 0.5894 & 0.4385 \\

20    
& \textbf{0.0167} & 0.0157 & \textbf{0.7982} & \textbf{0.3589}
& 0.0151 & \textbf{0.0152} & 0.7337 & 0.4017
& 0.0154 & 0.0161 & 0.6187 & 0.4396 \\

\midrule
\midrule

\textbf{PubMed}    &  &  &    &  &  &  &    &  &  &  &  &  \\
5    
& 0.0627 & 0.0527 & 0.6266 & 0.3844
& 0.0511 & \textbf{0.0416} & 0.6518 & 0.3739
& \textbf{0.0631} & 0.0487 & \textbf{0.7225} & \textbf{0.3676} \\

10   
& 0.0403 & 0.0401 & 0.6248 & 0.3989
& 0.0332 & 0.0333 & 0.6866 & 0.3949
& \textbf{0.0432} & \textbf{0.0321} & \textbf{0.7930} & \textbf{0.3384} \\

15   
& \textbf{0.0306} & 0.0287 & 0.6894 & 0.3895
& 0.0241 & 0.0232 & 0.7019 & 0.3981
& 0.0293 & \textbf{0.0227} & \textbf{0.7562} & \textbf{0.3699} \\

20   
& \textbf{0.0261} & 0.0246 & 0.7483 & 0.3763
& 0.0193 & 0.0184 & 0.7382 & 0.3922
& 0.0221 & \textbf{0.0155} & \textbf{0.8073} & \textbf{0.3458} \\

\midrule
\midrule

\textbf{Cybersecurity}    &      &      &      &      &      &      &      &      &      &      &      &      \\
5    
& 0.0423 & 0.0448 & 0.4917 & 0.4096
& \textbf{0.0437} & 0.0496 & 0.5619 & 0.4225
& 0.0409 & \textbf{0.0436} & \textbf{0.6504} & \textbf{0.4075} \\

10   
& \textbf{0.0305} & 0.0287 & 0.6155 & \textbf{0.4047}
& 0.0249 & 0.0275 & 0.6006 & 0.4425
& 0.0262 & \textbf{0.0257} & \textbf{0.6990} & 0.4080 \\

15   
& \textbf{0.0211} & 0.0212 & 0.5938 & 0.4201
& 0.0169 & 0.0217 & 0.5505 & 0.4517
& 0.0195 & \textbf{0.0193} & \textbf{0.7115} & \textbf{0.4147} \\

20   
& \textbf{0.0193} & 0.0190 & 0.6515 & 0.4100
& 0.0137 & 0.0164 & 0.6233 & 0.4546
& 0.0151 & \textbf{0.0152} & \textbf{0.7338} & \textbf{0.4017} \\

\bottomrule

\end{tabular}
\end{table*}

\begin{table*}[t]
\centering
\caption{Part B - Performance comparison of DeepSeek-R1-Distill-Llama-70B, DeepSeek-R1-Distill-Qwen-32B, and microsoft/phi-4 using BGE-m3 embeddings.}
\label{tab:link_prediction}
\begin{tabular}{lp{1cm}p{1cm}p{1cm}p{1cm}p{1cm}p{1cm}p{1cm}p{1cm}p{1cm}p{1cm}p{1cm}p{1cm}}
\toprule
& \multicolumn{4}{c}{\textbf{DeepSeek-R1-Distill-Llama-70B}} 
& \multicolumn{4}{c}{\textbf{DeepSeek-R1-Distill-Qwen-32B}} 
& \multicolumn{4}{c}{\textbf{microsoft/phi-4}}\\
\cmidrule(lr){2-5}\cmidrule(lr){6-9}\cmidrule(lr){10-13}
\textbf{Chunk Size} 
& {$\mathrm{\textbf{P}}_{\mathrm{mean}}$} & \textbf{$\mathrm{\textbf{P}}_{\mathrm{std}}$} 
& {$\mathrm{\textbf{P}\Omega}_{\mathrm{mean}}$} & {$\mathrm{\textbf{P}\Omega}_{\mathrm{std}}$}
& {$\mathrm{\textbf{P}}_{\mathrm{mean}}$} & \textbf{$\mathrm{\textbf{P}}_{\mathrm{std}}$} 
& {$\mathrm{\textbf{P}\Omega}_{\mathrm{mean}}$} & {$\mathrm{\textbf{P}\Omega}_{\mathrm{std}}$}
& {$\mathrm{\textbf{P}}_{\mathrm{mean}}$} & \textbf{$\mathrm{\textbf{P}}_{\mathrm{std}}$} 
& {$\mathrm{\textbf{P}\Omega}_{\mathrm{mean}}$} & {$\mathrm{\textbf{P}\Omega}_{\mathrm{std}}$} \\
\midrule

\textbf{Finance}    &  &  &    &  &  &  &    &  &  &  &  &  \\
5    
& \textbf{0.0449} & \textbf{0.0413} & 0.1476 & \textbf{0.1054}
& 0.0424 & 0.0459 & 0.2028 & 0.1248
& 0.0438 & 0.0443 & \textbf{0.2267} & 0.1349 \\

10   
& \textbf{0.0269} & \textbf{0.0235} & 0.0982 & \textbf{0.0667}
& 0.0265 & 0.0259 & 0.1210 & 0.0780
& \textbf{0.0269} & 0.0273 & \textbf{0.1418} & 0.0895 \\

15   
& 0.0180 & \textbf{0.0164} & 0.0799 & \textbf{0.0553}
& \textbf{0.0196} & 0.0193 & 0.0910 & 0.0605
& 0.0195 & 0.0203 & \textbf{0.1092} & 0.0687 \\

20   
& \textbf{0.0170} & 0.0166 & 0.0685 & \textbf{0.0472}
& 0.0152 & \textbf{0.0153} & 0.0772 & 0.0609
& 0.0155 & 0.0162 & \textbf{0.0868} & 0.0606 \\

\midrule
\midrule

\textbf{PubMed}    &  &  &    &  &  &  &    &  &  &  &  &  \\
5    
& \textbf{0.0667} & 0.0576 & 0.2790 & 0.1319
& 0.0537 & \textbf{0.0441} & 0.2473 & 0.1446
& 0.0647 & 0.0498 & \textbf{0.2818} & 0.1477 \\

10   
& 0.0414 & 0.0411 & 0.1807 & 0.1065
& 0.0339 & 0.0339 & 0.1499 & \textbf{0.0978}
& \textbf{0.0438} & \textbf{0.0325} & \textbf{0.1944} & 0.1143 \\

15   
& \textbf{0.0311} & 0.0291 & \textbf{0.1431} & 0.0974
& 0.0244 & 0.0236 & 0.1149 & 0.0774
& 0.0296 & \textbf{0.0231} & 0.1224 & \textbf{0.0651} \\

20   
& \textbf{0.0263} & 0.0248 & \textbf{0.1150} & 0.0747
& 0.0194 & 0.0186 & 0.1032 & 0.0852
& 0.0222 & \textbf{0.0156} & 0.1044 & \textbf{0.0596} \\

\midrule
\midrule

\textbf{Cybersecurity}    &      &      &      &      &      &      &      &      &      &      &      &      \\
5    
& 0.0445 & 0.0468 & 0.1959 & 0.1225
& \textbf{0.0449} & 0.0507 & 0.1964 & 0.1254
& 0.0422 & \textbf{0.0452} & \textbf{0.2028} & 0.1248 \\

10   
& \textbf{0.0311} & 0.0295 & 0.1243 & 0.0828
& 0.0253 & 0.0279 & \textbf{0.1278} & 0.1025
& 0.0265 & \textbf{0.0259} & 0.1210 & \textbf{0.0780} \\

15   
& \textbf{0.0213} & 0.0215 & 0.0972 & 0.0692
& 0.0170 & 0.0218 & \textbf{0.0982} & 0.0807
& 0.0196 & \textbf{0.0193} & 0.0910 & \textbf{0.0605} \\

20   
& \textbf{0.0194} & 0.0192 & \textbf{0.0822} & \textbf{0.0567}
& 0.0137 & 0.0164 & 0.0759 & 0.0569
& 0.0152 & \textbf{0.0153} & 0.0772 & 0.0609 \\

\bottomrule

\end{tabular}
\end{table*}

\begin{table*}[t]
\centering
\caption{Part A - Performance comparison of DeepSeek-R1-Distill-Llama-70B, DeepSeek-R1-Distill-Qwen-32B, and microsoft/phi-4 using Nomic embeddings.}
\label{tab:link_prediction}
\begin{tabular}{lp{1cm}p{1cm}p{1cm}p{1cm}p{1cm}p{1cm}p{1cm}p{1cm}p{1cm}p{1cm}p{1cm}p{1cm}}
\toprule
& \multicolumn{4}{c}{\textbf{DeepSeek-R1-Distill-Llama-70B}} 
& \multicolumn{4}{c}{\textbf{DeepSeek-R1-Distill-Qwen-32B}} 
& \multicolumn{4}{c}{\textbf{microsoft/phi-4}} \\
\cmidrule(lr){2-5}\cmidrule(lr){6-9}\cmidrule(lr){10-13}
\textbf{Chunk Size}
& {$\mathrm{\textbf{IOU}}_{\mathrm{mean}}$} & {$\mathrm{\textbf{IOU}}_{\mathrm{std}}$} & {$\mathrm{\textbf{Recall}}_{\mathrm{mean}}$} & {$\mathrm{\textbf{Recall}}_{\mathrm{std}}$}
& {$\mathrm{\textbf{IOU}}_{\mathrm{mean}}$} & {$\mathrm{\textbf{IOU}}_{\mathrm{std}}$} & {$\mathrm{\textbf{Recall}}_{\mathrm{mean}}$} & {$\mathrm{\textbf{Recall}}_{\mathrm{std}}$}
& {$\mathrm{\textbf{IOU}}_{\mathrm{mean}}$} & {$\mathrm{\textbf{IOU}}_{\mathrm{std}}$} & {$\mathrm{\textbf{Recall}}_{\mathrm{mean}}$} & {$\mathrm{\textbf{Recall}}_{\mathrm{std}}$} \\
\midrule

\textbf{Finance} & & & & & & & & & & & & \\

5
& 0.0386 & \textbf{0.0419} & 0.3921 & \textbf{0.4139}
& 0.0350 & 0.0422 & 0.4557 & 0.4313
& \textbf{0.0426} & 0.0470 & \textbf{0.4583} & 0.4239 \\

10
& 0.0167 & 0.0259 & 0.2592 & \textbf{0.3984}
& \textbf{0.0235} & \textbf{0.0255} & \textbf{0.5108} & 0.4427
& 0.0229 & 0.0296 & 0.4200 & 0.4369 \\

15
& 0.0083 & \textbf{0.0156} & 0.1977 & \textbf{0.3550}
& \textbf{0.0138} & 0.0189 & \textbf{0.4039} & 0.4523
& 0.0124 & 0.0203 & 0.3207 & 0.4233 \\

20
& \textbf{0.0102} & 0.0183 & 0.2761 & \textbf{0.4144}
& 0.0078 & \textbf{0.0124} & \textbf{0.3226} & 0.4372
& 0.0084 & 0.0142 & 0.2980 & 0.4192 \\

\midrule
\midrule

\textbf{PubMed} & & & & & & & & & & & & \\

5
& 0.0571 & 0.0597 & \textbf{0.5031} & 0.4160
& 0.0482 & 0.0496 & 0.4793 & 0.4061
& 0.0350 & 0.0422 & 0.4557 & 0.4313 \\

10
& 0.0332 & 0.0387 & 0.4956 & 0.4316
& 0.0312 & 0.0332 & \textbf{0.5178} & 0.4220
& 0.0235 & 0.0255 & 0.5108 & 0.4427 \\

15
& \textbf{0.0310} & 0.0290 & 0.1430 & 0.0973
& 0.0231 & 0.0297 & \textbf{0.4946} & 0.4492
& 0.0138 & 0.0189 & 0.4039 & 0.4523 \\

20
& \textbf{0.0262} & 0.0247 & 0.1149 & 0.0746
& 0.0158 & 0.0200 & \textbf{0.4686} & 0.4606
& 0.0078 & 0.0124 & 0.3226 & 0.4372 \\

\midrule
\midrule

\textbf{Cybersecurity} & & & & & & & & & & & & \\

5
& 0.0422 & 0.0482 & 0.3829 & 0.3929
& \textbf{0.0432} & 0.0565 & 0.4317 & 0.4157
& 0.0350 & \textbf{0.0422} & \textbf{0.4557} & 0.4313 \\

10
& 0.0224 & 0.0304 & 0.3588 & 0.4160
& 0.0228 & 0.0307 & 0.4142 & 0.4384
& \textbf{0.0235} & \textbf{0.0255} & \textbf{0.5108} & 0.4427 \\

15
& \textbf{0.0155} & 0.0234 & \textbf{0.3385} & 0.4212
& 0.0126 & 0.0207 & 0.3074 & 0.4095
& 0.0138 & \textbf{0.0189} & 0.4039 & \textbf{0.4523} \\

20
& \textbf{0.0105} & 0.0167 & \textbf{0.2797} & 0.3881
& 0.0100 & 0.0170 & 0.2782 & 0.3977
& 0.0078 & \textbf{0.0124} & 0.3226 & \textbf{0.4372} \\

\bottomrule
\end{tabular}
\end{table*}

\begin{table*}[t]
\centering
\caption{Part B - Performance comparison of DeepSeek-R1-Distill-Llama-70B, DeepSeek-R1-Distill-Qwen-32B, and microsoft/phi-4 using Nomic embeddings.}
\label{tab:link_prediction}
\begin{tabular}{lp{1cm}p{1cm}p{1cm}p{1cm}p{1cm}p{1cm}p{1cm}p{1cm}p{1cm}p{1cm}p{1cm}p{1cm}}
\toprule
& \multicolumn{4}{c}{\textbf{DeepSeek-R1-Distill-Llama-70B}} 
& \multicolumn{4}{c}{\textbf{DeepSeek-R1-Distill-Qwen-32B}} 
& \multicolumn{4}{c}{\textbf{microsoft/phi-4}}\\
\cmidrule(lr){2-5}\cmidrule(lr){6-9}\cmidrule(lr){10-13}
\textbf{Chunk Size} 
& {$\mathrm{\textbf{P}}_{\mathrm{mean}}$} & \textbf{$\mathrm{\textbf{P}}_{\mathrm{std}}$} 
& {$\mathrm{\textbf{P}\Omega}_{\mathrm{mean}}$} & {$\mathrm{\textbf{P}\Omega}_{\mathrm{std}}$}
& {$\mathrm{\textbf{P}}_{\mathrm{mean}}$} & \textbf{$\mathrm{\textbf{P}}_{\mathrm{std}}$} 
& {$\mathrm{\textbf{P}\Omega}_{\mathrm{mean}}$} & {$\mathrm{\textbf{P}\Omega}_{\mathrm{std}}$}
& {$\mathrm{\textbf{P}}_{\mathrm{mean}}$} & \textbf{$\mathrm{\textbf{P}}_{\mathrm{std}}$} 
& {$\mathrm{\textbf{P}\Omega}_{\mathrm{mean}}$} & {$\mathrm{\textbf{P}\Omega}_{\mathrm{std}}$} \\
\midrule

\textbf{Finance} & & & & & & & & & & & & \\

5
& 0.0410 & 0.0440 & 0.1476 & \textbf{0.1054}
& 0.0363 & \textbf{0.0439} & 0.2028 & 0.1248
& \textbf{0.0445} & 0.0494 & \textbf{0.2267} & 0.1349 \\

10
& 0.0171 & \textbf{0.0262} & 0.0982 & \textbf{0.0667}
& \textbf{0.0241} & \textbf{0.0262} & 0.1210 & 0.0780
& 0.0234 & 0.0301 & \textbf{0.1418} & 0.0895 \\

15
& 0.0085 & \textbf{0.0158} & 0.0799 & \textbf{0.0553}
& \textbf{0.0140} & 0.0191 & 0.0910 & 0.0605
& 0.0127 & 0.0206 & \textbf{0.1092} & 0.0687 \\

20
& \textbf{0.0105} & 0.0192 & 0.0685 & \textbf{0.0472}
& 0.0078 & \textbf{0.0126} & 0.0772 & 0.0609
& 0.0085 & 0.0143 & \textbf{0.0868} & 0.0606 \\

\midrule
\midrule

\textbf{PubMed} & & & & & & & & & & & & \\

5
& \textbf{0.0605} & 0.0622 & \textbf{0.2790} & 0.1319
& 0.0511 & 0.0544 & 0.2473 & 0.1446
& 0.0363 & \textbf{0.0439} & 0.2028 & \textbf{0.1248} \\

10
& \textbf{0.0343} & 0.0397 & \textbf{0.1807} & 0.1065
& 0.0320 & 0.0345 & 0.1499 & 0.0978
& 0.0241 & \textbf{0.0262} & 0.1210 & \textbf{0.0780} \\

15
& \textbf{0.0264} & 0.0318 & \textbf{0.1431} & 0.0974
& 0.0234 & 0.0302 & 0.1149 & 0.0774
& 0.0140 & \textbf{0.0191} & 0.0910 & \textbf{0.0605} \\

20
& \textbf{0.0190} & 0.0255 & \textbf{0.1150} & 0.0747
& 0.0160 & 0.0202 & 0.1032 & 0.0852
& 0.0078 & \textbf{0.0126} & 0.0772 & \textbf{0.0609} \\

\midrule
\midrule

\textbf{Cybersecurity} & & & & & & & & & & & & \\

5
& \textbf{0.0446} & 0.0499 & 0.1959 & \textbf{0.1225}
& \textbf{0.0446} & 0.0579 & 0.1964 & 0.1254
& 0.0363 & \textbf{0.0439} & \textbf{0.2028} & 0.1248 \\

10
& 0.0229 & 0.0309 & 0.1243 & 0.0828
& 0.0231 & 0.0312 & \textbf{0.1278} & 0.1025
& \textbf{0.0241} & \textbf{0.0262} & 0.1210 & \textbf{0.0780} \\

15
& \textbf{0.0157} & 0.0236 & 0.0972 & 0.0692
& 0.0128 & 0.0209 & \textbf{0.0982} & 0.0807
& 0.0140 & \textbf{0.0191} & 0.0910 & \textbf{0.0605} \\

20
& \textbf{0.0106} & 0.0168 & \textbf{0.0822} & \textbf{0.0567}
& 0.0101 & 0.0173 & 0.0759 & 0.0569
& 0.0078 & \textbf{0.0126} & 0.0772 & 0.0609 \\

\bottomrule
\end{tabular}
\end{table*}

\begin{table*}[t]
\centering
\caption{Part A - Performance comparison of DeepSeek-R1-Distill-Llama-70B, DeepSeek-R1-Distill-Qwen-32B, and microsoft/phi-4 using ALL-MINILM Embeddings.}
\label{tab:link_prediction}
\begin{tabular}{lp{1cm}p{1cm}p{1cm}p{1cm}p{1cm}p{1cm}p{1cm}p{1cm}p{1cm}p{1cm}p{1cm}p{1cm}}
\toprule
& \multicolumn{4}{c}{\textbf{DeepSeek-R1-Distill-Llama-70B}} 
& \multicolumn{4}{c}{\textbf{DeepSeek-R1-Distill-Qwen-32B}} 
& \multicolumn{4}{c}{\textbf{microsoft/phi-4}}\\
\cmidrule(lr){2-5}\cmidrule(lr){6-9}\cmidrule(lr){10-13}
\textbf{Chunk Size}
& {$\mathrm{\textbf{IOU}}_{\mathrm{mean}}$} & {$\mathrm{\textbf{IOU}}_{\mathrm{std}}$} & {$\mathrm{\textbf{Recall}}_{\mathrm{mean}}$} & {$\mathrm{\textbf{Recall}}_{\mathrm{std}}$}
& {$\mathrm{\textbf{IOU}}_{\mathrm{mean}}$} & {$\mathrm{\textbf{IOU}}_{\mathrm{std}}$} & {$\mathrm{\textbf{Recall}}_{\mathrm{mean}}$} & {$\mathrm{\textbf{Recall}}_{\mathrm{std}}$}
& {$\mathrm{\textbf{IOU}}_{\mathrm{mean}}$} & {$\mathrm{\textbf{IOU}}_{\mathrm{std}}$} & {$\mathrm{\textbf{Recall}}_{\mathrm{mean}}$} & {$\mathrm{\textbf{Recall}}_{\mathrm{std}}$} \\
\midrule

\textbf{Finance} &&&&&&&&&&& \\
5  & 0.0337 & \textbf{0.0439} & 0.3543 & \textbf{0.3968}
   & 0.0301 & 0.0440 & 0.4144 & 0.4405
   & \textbf{0.0375} & 0.0449 & \textbf{0.4309} & 0.4301 \\

10 & \textbf{0.0221} & 0.0279 & 0.4172 & \textbf{0.4230}
   & 0.0162 & \textbf{0.0256} & 0.4148 & 0.4563
   & 0.0203 & 0.0261 & \textbf{0.4275} & 0.4449 \\

15 & \textbf{0.0140} & 0.0193 & 0.3822 & \textbf{0.4242}
   & 0.0115 & 0.0195 & \textbf{0.4029} & 0.4624
   & 0.0137 & \textbf{0.0191} & 0.3955 & 0.4461 \\

20 & \textbf{0.0110} & 0.0160 & 0.3644 & \textbf{0.4282}
   & 0.0075 & \textbf{0.0133} & 0.3533 & 0.4553
   & 0.0098 & 0.0152 & \textbf{0.3774} & 0.4515 \\

\midrule
\midrule

\textbf{PubMed} &&&&&&&&&&& \\
5  & 0.0434 & 0.0574 & 0.3976 & 0.4235
   & 0.0428 & \textbf{0.0474} & 0.4687 & 0.4237
   & \textbf{0.0567} & 0.0527 & \textbf{0.5652} & \textbf{0.4200} \\

10 & 0.0344 & 0.0422 & 0.5154 & 0.4336
   & 0.0295 & 0.0371 & 0.5092 & 0.4592
   & \textbf{0.0371} & \textbf{0.0289} & \textbf{0.6660} & \textbf{0.4222} \\

15 & 0.0212 & 0.0329 & 0.4428 & 0.4593
   & 0.0218 & 0.0252 & \textbf{0.5468} & 0.4416
   & \textbf{0.0225} & \textbf{0.0227} & 0.5126 & \textbf{0.4391} \\

20 & 0.0148 & 0.0212 & 0.4258 & \textbf{0.4523}
   & 0.0176 & 0.0213 & \textbf{0.5862} & 0.4638
   & \textbf{0.0182} & \textbf{0.0171} & 0.5815 & 0.4552 \\

\midrule
\midrule

\textbf{Cybersecurity} &&&&&&&&&&& \\
5  & 0.0351 & 0.0485 & 0.3427 & \textbf{0.3835}
   & \textbf{0.0394} & 0.0499 & 0.4605 & 0.4317
   & 0.0322 & \textbf{0.0374} & \textbf{0.4629} & 0.4323 \\

10 & \textbf{0.0249} & 0.0290 & 0.4189 & \textbf{0.4033}
   & 0.0239 & 0.0280 & 0.5147 & 0.4440
   & 0.0225 & \textbf{0.0259} & \textbf{0.5512} & 0.4567 \\

15 & \textbf{0.0185} & 0.0240 & 0.4237 & \textbf{0.4346}
   & 0.0119 & \textbf{0.0168} & 0.3974 & 0.4463
   & 0.0148 & 0.0175 & \textbf{0.5306} & 0.4522 \\

20 & \textbf{0.0155} & 0.0179 & 0.4678 & \textbf{0.4414}
   & 0.0103 & \textbf{0.0139} & 0.4318 & 0.4639
   & 0.0116 & 0.0153 & \textbf{0.4992} & 0.4602 \\

\bottomrule
\end{tabular}
\end{table*}

\begin{table*}[t]
\centering
\caption{Part B - Performance comparison of DeepSeek-R1-Distill-Llama-70B, DeepSeek-R1-Distill-Qwen-32B, and microsoft/phi-4 using ALL-MINILM Embeddings.}
\label{tab:link_prediction}
\begin{tabular}{lp{1cm}p{1cm}p{1cm}p{1cm}p{1cm}p{1cm}p{1cm}p{1cm}p{1cm}p{1cm}p{1cm}p{1cm}}
\toprule
& \multicolumn{4}{c}{\textbf{DeepSeek-R1-Distill-Llama-70B}} 
& \multicolumn{4}{c}{\textbf{DeepSeek-R1-Distill-Qwen-32B}} 
& \multicolumn{4}{c}{\textbf{microsoft/phi-4}}\\
\cmidrule(lr){2-5}\cmidrule(lr){6-9}\cmidrule(lr){10-13}
\textbf{Chunk Size} 
& {$\mathrm{\textbf{P}}_{\mathrm{mean}}$} & \textbf{$\mathrm{\textbf{P}}_{\mathrm{std}}$} 
& {$\mathrm{\textbf{P}\Omega}_{\mathrm{mean}}$} & {$\mathrm{\textbf{P}\Omega}_{\mathrm{std}}$}
& {$\mathrm{\textbf{P}}_{\mathrm{mean}}$} & \textbf{$\mathrm{\textbf{P}}_{\mathrm{std}}$} 
& {$\mathrm{\textbf{P}\Omega}_{\mathrm{mean}}$} & {$\mathrm{\textbf{P}\Omega}_{\mathrm{std}}$}
& {$\mathrm{\textbf{P}}_{\mathrm{mean}}$} & \textbf{$\mathrm{\textbf{P}}_{\mathrm{std}}$} 
& {$\mathrm{\textbf{P}\Omega}_{\mathrm{mean}}$} & {$\mathrm{\textbf{P}\Omega}_{\mathrm{std}}$} \\
\midrule

\textbf{Finance} &&&&&&&&&&& \\
5  & 0.0357 & 0.0465 & 0.2054 & \textbf{0.1217}
   & 0.0312 & \textbf{0.0458} & 0.2048 & 0.1372
   & \textbf{0.0392} & 0.0470 & \textbf{0.2267} & 0.1349 \\

10 & \textbf{0.0227} & 0.0289 & 0.1351 & \textbf{0.0817}
   & 0.0165 & \textbf{0.0259} & 0.1275 & 0.0937
   & 0.0207 & 0.0267 & \textbf{0.1418} & 0.0895 \\

15 & \textbf{0.0142} & 0.0197 & 0.1038 & \textbf{0.0665}
   & 0.0116 & 0.0199 & 0.0951 & 0.0724
   & 0.0138 & \textbf{0.0194} & \textbf{0.1092} & 0.0687 \\

20 & \textbf{0.0111} & 0.0163 & \textbf{0.0871} & \textbf{0.0563}
   & 0.0076 & \textbf{0.0134} & 0.0766 & 0.0605
   & 0.0099 & 0.0153 & 0.0868 & 0.0606 \\

\midrule
\midrule

\textbf{PubMed} &&&&&&&&&&& \\
5  & 0.0461 & 0.0607 & 0.2790 & \textbf{0.1319}
   & 0.0452 & \textbf{0.0513} & 0.2473 & 0.1446
   & \textbf{0.0589} & 0.0546 & \textbf{0.2818} & 0.1477 \\

10 & 0.0354 & 0.0435 & 0.1807 & 0.1065
   & 0.0301 & 0.0381 & 0.1499 & \textbf{0.0978}
   & \textbf{0.0377} & \textbf{0.0290} & \textbf{0.1944} & 0.1143 \\

15 & 0.0215 & 0.0333 & \textbf{0.1431} & 0.0974
   & 0.0222 & 0.0257 & 0.1149 & 0.0774
   & \textbf{0.0228} & \textbf{0.0231} & 0.1224 & \textbf{0.0651} \\

20 & 0.0150 & 0.0215 & \textbf{0.1150} & 0.0747
   & 0.0177 & 0.0214 & 0.1032 & 0.0852
   & \textbf{0.0183} & \textbf{0.0172} & 0.1044 & \textbf{0.0596} \\

\midrule
\midrule

\textbf{Cybersecurity} &&&&&&&&&&& \\
5  & 0.0370 & 0.0503 & 0.1959 & \textbf{0.1225}
   & \textbf{0.0407} & 0.0508 & 0.1964 & 0.1254
   & 0.0338 & \textbf{0.0393} & \textbf{0.2028} & 0.1248 \\

10 & \textbf{0.0255} & 0.0295 & 0.1243 & 0.0828
   & 0.0244 & 0.0284 & \textbf{0.1278} & 0.1025
   & 0.0229 & \textbf{0.0263} & 0.1210 & \textbf{0.0780} \\

15 & \textbf{0.0187} & 0.0242 & 0.0972 & 0.0692
   & 0.0120 & \textbf{0.0170} & \textbf{0.0982} & 0.0807
   & 0.0150 & 0.0178 & 0.0910 & \textbf{0.0605} \\

20 & \textbf{0.0156} & 0.0181 & \textbf{0.0822} & \textbf{0.0567}
   & 0.0104 & \textbf{0.0140} & 0.0759 & 0.0569
   & 0.0117 & 0.0154 & 0.0772 & 0.0609 \\

\bottomrule
\end{tabular}
\end{table*}

\section{Conclusion}

In this work, we presented a comprehensive evaluation of \textbf{DeepSeek-R1} across multiple domains (Finance, PubMed, and Cybersecurity) and varied chunk sizes, employing three categories of embeddings (BGE-M3, Nomic, and ALL-MINILM) as well as three distilled large language models (\textbf{DeepSeek-R1-Distill-Llama-70B}, \textbf{DeepSeek-R1-Distill-Qwen-32B}, and \textbf{microsoft/phi-4}). Our findings reveal consistent trade-offs between precision-oriented and recall-oriented performance. In particular, smaller chunk sizes generally enhance \(\mathrm{IOU}\), \(\mathrm{P}\), and \(\mathrm{P}\Omega\), capturing fewer irrelevant spans and thereby boosting precision. Larger chunk sizes, on the other hand, improve \(\mathrm{Recall}\), covering broader portions of text and retrieving more relevant information at the expense of precision.

Among the models evaluated, \textbf{DeepSeek-R1-Distill-Llama-70B} often achieves competitive coverage (as indicated by \(\mathrm{IOU}\)) and stable recall results, while \textbf{microsoft/phi-4} tends to excel in achieving high recall and \(\mathrm{P}\Omega\). \textbf{DeepSeek-R1-Distill-Qwen-32B} stands out for its relatively low variance in both \(\mathrm{IOU}\) and precision. Consequently, no single model completely dominates across all evaluation metrics and domains, underscoring the necessity of aligning model selection and chunk-size configuration with task-specific objectives (e.g., higher recall \textit{vs.} greater precision).

Overall, our experiments highlight the importance of tailoring both the retrieval model and chunk-size strategy to domain-specific needs. Future work may explore extending \textbf{DeepSeek-R1} to larger or more specialized corpora, improving the efficiency of index construction for very large-scale collections, and investigating ensemble or hybrid approaches that combine the complementary strengths of multiple embeddings and models to further optimize retrieval performance.

\bibliographystyle{IEEEtran}
\nocite{*}
\bibliography{references}
\vspace{12pt}
\end{document}